\documentclass[a4paper,twoside]{article}

\usepackage{epsfig}
\usepackage{subfigure}
\usepackage{calc}
\usepackage{amssymb}
\usepackage{amstext}
\usepackage{amsmath}
\usepackage{amsthm}
\usepackage{multicol}
\usepackage{pslatex}
\usepackage{apalike}
\usepackage{SciTePress}
\usepackage[small]{caption}
\usepackage{booktabs}

\begin{document}

\title{Visual Tracking with Similarity Matching Ratio}

\author{\authorname{Aysegul Dundar\sup{1}, Jonghoon Jin\sup{2} and Eugenio Culurciello\sup{1}}
\affiliation{\sup{1}Weldon School of Biomedical Engineering, Purdue University, West Lafayette, IN, USA}
\affiliation{\sup{2}Electrical and Computer Engineering, Purdue University, West Lafayette, IN, USA}
\email{\{adundar, jhjin, euge\}@purdue.edu}
}

\keywords{Tracking; SMR; Similarity Matching Ratio;Template Matching.}

\abstract{\small This paper presents a novel approach to visual tracking: Similarity Matching Ratio (SMR). The traditional approach of tracking is minimizing some measures of the difference between the template and a patch from the frame. This approach is vulnerable to outliers and drastic appearance changes and an extensive study is focusing on making the approach more tolerant to them. However, this often results in longer, corrective algorithms which do not solve the original problem. This paper proposes a novel approach to the definition of the tracking problems, SMR, which turns the differences into a probability measure. Only pixel differences below a threshold count towards deciding the match, the rest are ignored. This approach makes the SMR tracker robust to outliers and points that dramaticaly change appearance. The SMR tracker is tested on challenging video sequences and achieved state-of-the-art performance.}

\onecolumn \maketitle \normalsize \vfill

\section{\uppercase{Introduction}}
\label{sec:introduction}

\noindent Visual tracking of objects in a scene is a very important component of a unified robotic vision system. Robots need to track objects in order to interact. As such as they move closer, robots and other autonomous vehicles will have to avoid other moving objects, humans, animals, as they operate in our everyday environment.

The human visual system object tracking performance is currently unsurpassed by engineered systems, thus our research tries to take inspiration and reverse-engineer the known principles of cortical processing during visual tracking. Visual tracking is a complex task, with neuroscience studies of cortical processing painting an incomplete picture, and thus is only partially able to guide the design of a synthetic solution. Nevertheless a few key features arise from studying the human visual system and its tacking abilities: (1) the human visual system is not limited to three-dimensional conventional objects in space, rather is able to track a set of visual  features  \cite{Blaser2000}. Thus ÒobjectÓ in this paper refers to a distinct group of features in the two-dimensional space. (2) It is not necessary for humans to have knowledge of the object class before visual tracking, and (3) humans can track an object after a very brief presentation. Even though the human visual system does not operate with ÒframesÓ it is common to desire synthetic systems to be able to track from a single frame, or just a few (tens).

{ \begin{table*}\small
\caption{Properties of the video dataset used in this work \cite{KalalTLD2010}. }\label{tab:properties} \centering
\begin{tabular}{lcccccc}
 \toprule 
 \multicolumn{4}{c}{}{Video Sequence}  \\
 \cmidrule{2-7}
  &1. David  & 2. Jumping & 3. Pedestrian1 &  4. Pedestrian2 &  5.  Pedestrian3 &  6. Car\\
  \midrule
  Number of Frames	& 761	& 313	& 140 	& 338	& 184	& 945\\
    \midrule
  Camera Movement		& yes		& yes 		& yes 		&yes		& yes		& yes\\
      \midrule
 Partial Occlusion		& yes		& no 			& no 			&yes		& yes		& yes\\
      \midrule
  Full Occlusion 		& no			& no 			& no 			&yes		& yes		& yes\\
      \midrule
  Pose Change      		& yes		& no 			& no			&no		& no			& no\\
      \midrule
  Illumination Change	& yes		& no 			& no 			&no		& no			& no\\
      \midrule
   Scale change			& yes		& no 			& no			&no		& no			& no\\
      \midrule
   Similar Objects		& no			& no			& no 			&yes		& yes		& yes\\
   \bottomrule
   
\end{tabular}
\end{table*}

{ \begin{table*}\small
\caption{Number of correctly tracked frames from the state-of-art trackers and the SMR tracker. Table is taken and modified from \cite{Kalal2010}. }\label{tab:example1} \centering
\begin{tabular}{lcccccc}
 \toprule 
 \multicolumn{4}{c}{}{Video Sequence}  \\
 \cmidrule{2-7}
  &1. David  & 2. Jumping & 3. Pedestrian1 &  4. Pedestrian2 &  5.  Pedestrian3 &  6. Car\\
  \midrule
  Number of Frames	& 761	& 313	& 140 	& 338	& 184	& 945\\
    \midrule

   \cite{Lim2004}	& 17		& 75 		& 11 		& 33		& 50		& 163\\
   
  \cite{Collins2005}	& n/a		& 313	& 6  		& 8 		& 5		& n/a\\
   
  \cite{Avidan2007}	& 94		& 44		& 22 		& 118	& 53		& 10\\
   
  \cite{Babenko2009}& 135	& 313	& 101 	& 37		& 49		& 45\\
   
  \cite{Kalal2010}	& 761	& 170	& 140 	& 97		& 52		& 510\\
    \midrule   
      
   SMR (this work)			& 761	& 313	& 140 	& 236	& 66		& 510\\
   \bottomrule
   
\end{tabular}
\end{table*}

Visual tracking in artificial systems has been studied for decades, with laudable results \cite{Yilmaz2006}. In this paper we focus on bio-inspired visual tracking systems that can be part of a unified neurally-inspired vision system. Ideally, a unified visual model would be able to parse and detect an object every ÒframeÓ, but right now there is no bio-inspired model that can do this in real-time \cite{DiCarlo2012,LeCun2004,Serre2007}. Deep neural networks come close to this performance when trained to look for a single object on a large collection of images \cite{Sermanet2011}. 

When we think of visual tracking we often have in mind a familiar object in space. But humans are able to track any localized variation in a 2D field, such as a set of features \cite{Blaser2000}. It is a high-SNR peak-detector that allows us to track a puff of smoke or a cloud, for example. A bio-inspired synthetic visual tracker is generally thought of having two outputs of the same unified stream: one is a deep neural network classifier that is capable of categorizing object, another is a shallower classifier that can group features into ÒobjectnessÓ. The first deep system is used to be able to continue tracking an object as it disappears and reappears in the scene, while the second system provides rapid grouping of local features, by tracking local maxima in the retinal space. Such distinction might be necessary as a deep system will need 100-200ms to process one visual scene \cite{Thorpe1996}, while tracking without predicting object movement, as the one required for the oculo-motor control of smooth-pursuit \cite{Wilmer2007}, requires faster processing of the visual stream.

Inspired by recent findings on shallow feature extractors of the visual cortex \cite{Vintch2010}, we postulate that simple tracking processes are based on a shallow neural network that can identify quickly similarities between object features repeated in time. We propose an algorithm that can track and extract motion of an object based on the similarity between local features observed in subsequent frames. The local features are initially defined as a bounding box that defines the object to track.

Traditional template matching algorithms define the tracking problem as follows: we are given two images $F(x,y)$ and $G(x,y)$ which represent the pixels values at each location $(x,y)$. We want to find the distance vector $(h_{1},h_{2})$ that minimizes some measures of the difference between $F(x+h_{1},y+h_{2})$ and $G(x,y)$ \cite{Lucas1981}. The measures can be cross correlation, image intensity, color features, image gradients or color histograms. However, this traditional definition of tracking suffers from outliers or regions that drastically change their appearance or disappear from the scene. 

In our work we change this definition of tracking and propose a novel approach, Similarity Match Ratio (SMR). Instead of trying to minimize some measures of difference between $F(x+h_{1},y+h_{2})$ and $G(x,y)$, we want to find $(h_{1},h_{2})$ that gives the best match ratio between $F(x+h_{1},y+h_{2})$ and $G(x,y)$. To do this, we are turning differences into a probability value and accumulating them for every pixel that has a good match. If there is no good match between $F(x+h_{1},y+h_{2})$ and $G(x,y)$, the difference gives zero probability because we are not interested in how badly the two pixels match. This approach is more robust to appearance change, disappearance and outliers. The method is tested on challenging benchmark video sequences which include camera movement, partial/full occlusion, illuminance change, scale change and similar objects. State-of-the-art performance is achieved from these video sequences.

\section{\uppercase{Previous Work}}

Most popular trackers that are based on the traditional definition of the tracking problem (e.g. Sum-of-Squared-Distances (SSD), Sum-of-Absolute-Differences (SAD), Lucas-Kanade tracker) try to find distance vector $(h_{1},h_{2})$ that minimizes the difference between $F(x+h_{1},y+h_{2})$ and $G(x,y)$ either on the grayscale or color image. However, the template $G(x,y)$ may be including outliers or some parts that dramatically change or disappear, which cause tracking failure. The common approach to overcome these tracking failures is that trackers should not treat all pixels in a uniform manner but eliminate outliers from the computation.

Some studies \cite{Comaniciu2003,Shi1994} proposed using a weighted histogram as a measure to minimize for the tracking. By assuming that pixels close to the center are the most reliable, these methods weigh them higher, since occlusions and interferences tend to occur close to boundaries. However, a dramatical change in the appearance can occur even in the center, which cannot be handled by this method.

There are studies that aim to detect outliers and suppress them from the computation. \cite{Hager1998} uses the common approach that outliers produce large image differences that can be detected by the estimation process \cite{Black1998}. Residuals are calculated iteratively and if the variations of the residual are bigger than a user defined threshold they are considered outliers and suppressed. \cite{Ishikawa2002} uses the spatial coherence property of the outliers which means that outliers tend to form a spatially coherent group rather than being randomly distributed across the template. In that work the template is divided into blocks and constant weights are assigned for each block. If the image differences of the blocks between the frames are large, it means these blocks include a significant amount of outliers. The method excludes the blocks that contain outliers from the computation of minimization. These methods are robust to outliers. However, they are computationally expensive. 

\cite{Kalal2010}  tracks the points from the template back and forth between the previous frame and current frame and validates the detection. This method enables trackers to avoid tracking points that disappear from the camera view or change appearance drastically. Before our work, Kalal's tracker was the state-of-the-art.

\begin{figure}[!h]
  \centering
   {\epsfig{file = 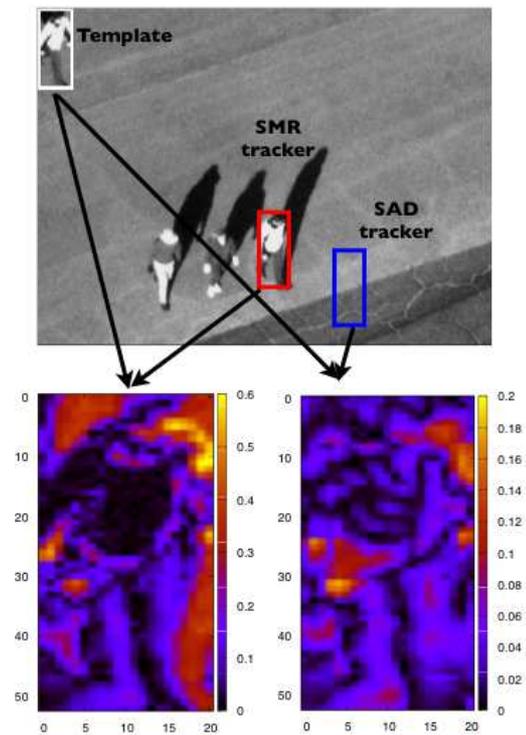, trim =9.5cm 3cm 4cm 5cm, clip ,  width = 12cm}}
  \caption{(Top) The red box is the SMR tracker's output, the blue box is the SAD tracker's output. The ground-truth from the first frame is used as a template which is shown on the left top corner of the frame. (Bottom) The absolute differences for each pixel between the template and result from the SMR tracker are mapped on the left and from the SAD tracker on the right. Dark values (close to zero) report a better match. Note that even though there are higher differences, the SMR tracker is able to find the correct patch. }
  \label{fig:example1}
 \end{figure}

 \begin{figure*}[ht]
\centering

\subfigure[]{
    \epsfig{file = 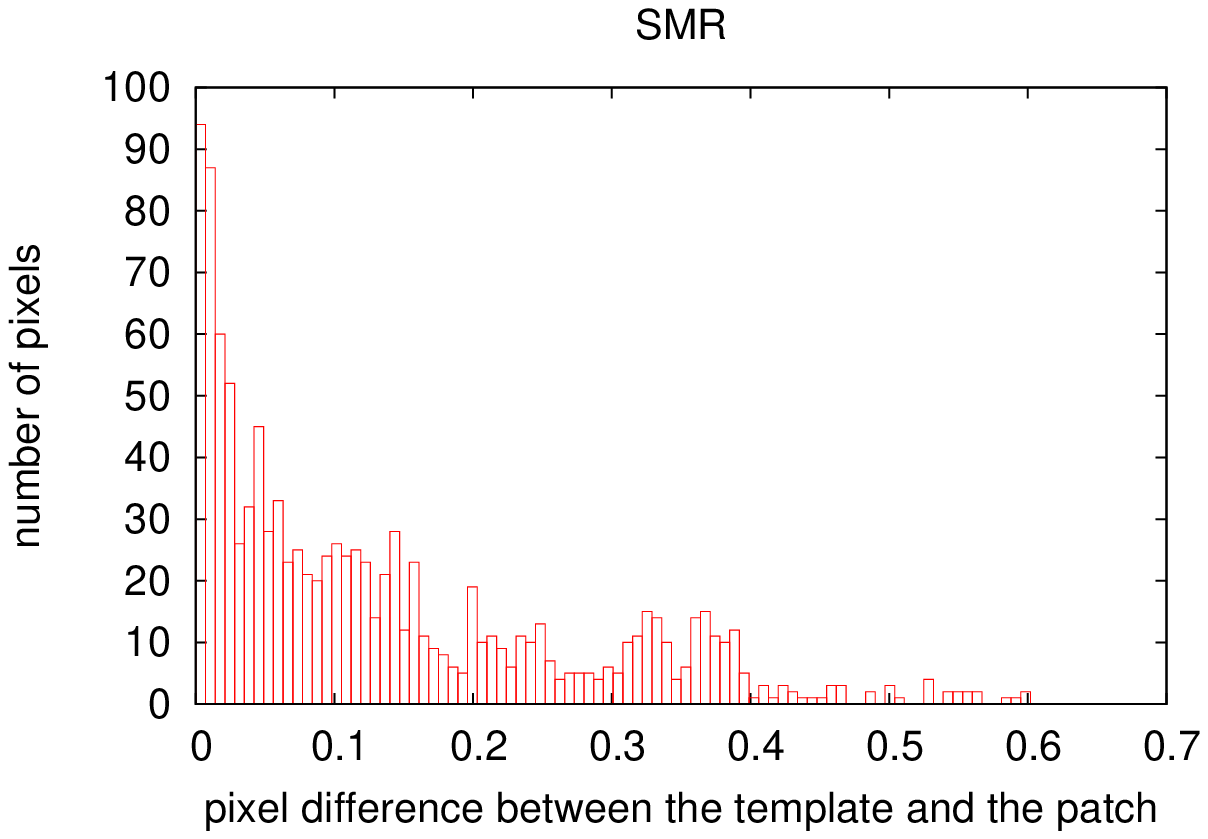, width =7cm}
    \label{fig:SMRHist}
}
\subfigure[]{
    \epsfig{file = 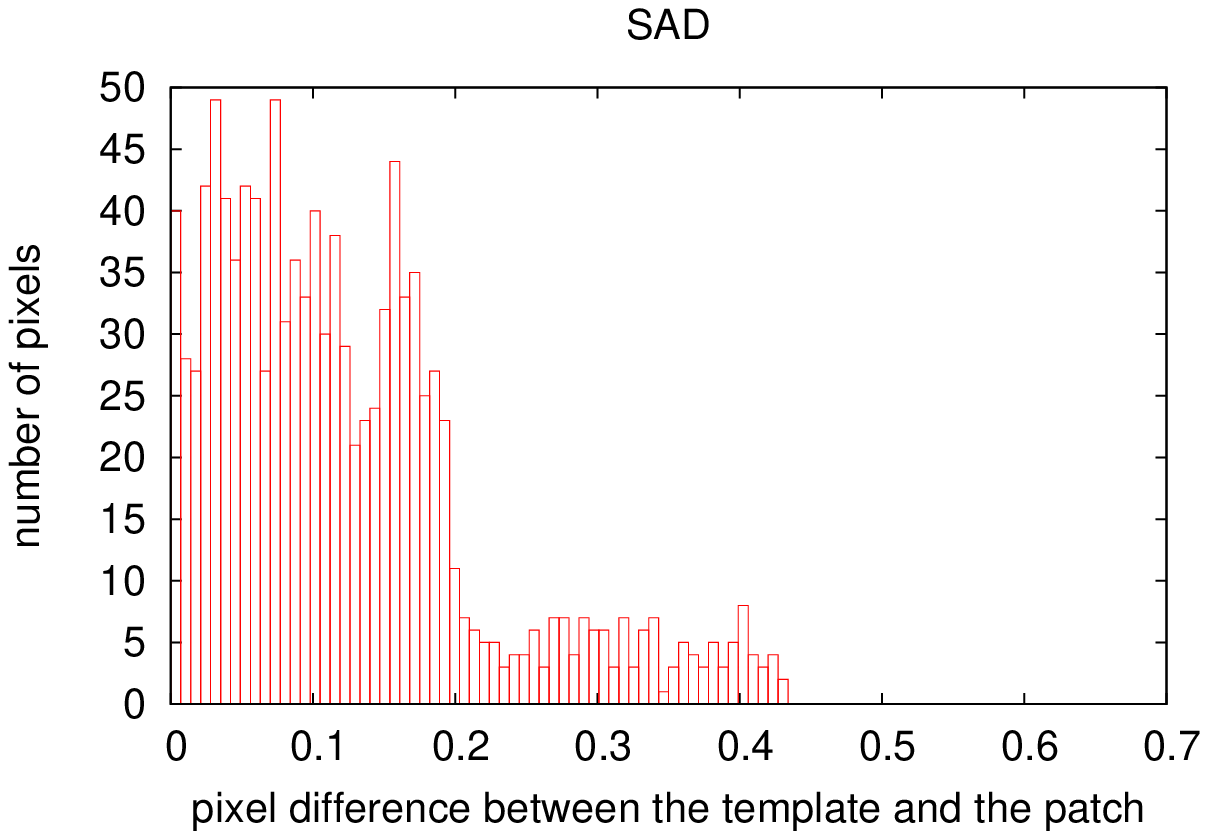, width = 7cm}
     \label{fig:SADHist}  
}
\caption[Optional caption for list of figures]{Histogram of the pixel differences that were mapped in Figure \ref{fig:example1}. (a) Map between the template and result from the SMR tracker and  (b) result from the SAD tracker. The SAD tracker minimizes  the number pixels with large differences, whereas the SMR tracker maximizes the number of pixels that have small differences. }
\label{fig:Hist}
\end{figure*}

\section{\uppercase{Similarity Matching Ratio (SMR) Tracker}}

The SMR tracker uses a modified template-matching algorithm. In this algorithm, we look for similarity between a template $G(x,y)$ and patches of a new video frame $F(x+h_{1},y+h_{2})$. The SMR computes the difference between the template and the patches at each pixel. Templates are moved convolutionally on the new video frame, and stepped by one pixel. If this difference is lower than a threshold, it is summed to the output after negative exponential distance conversion. This thresholding eliminates outlying pixels, in such a way that they do not appear in the final output. The SMR algorithm is as follows:
 
\begin{enumerate}
      \item The search area, $(h_{1}, h_{2})$, is limited to the neighborhood of the target's previous position.
       \item For each pixel in the template $G(x,y)$, the method is checking if the condition $F(x+h_{1},y+h_{2})-G(x,y)\leq\mathrm{\alpha}$ is satisfied, where $\alpha$ is a dynamic threshold defined in 6.
       \item If satisfied, we are interested in how close the match is, so the pixel difference is converted into a probability value $p$ by $p = \exp(-|F(x+h_{1},y+h_{2})-G(x,y)|)$. If not these pixels are ignored.
       \item The probability values are summed up for each patch. The algorithm finds the  $(h_{1},h_{2})$  that gives the highest similarity matching ratio, $\arg\max_{h_1, h_2}\sum p$.
       \item     $G(x,y)_{t+1}=F(x+h_{1},y+h_{2})_{t}$ The patch is extracted in every detection and assigned as new template.  
       \item  Dynamic threshold ${\alpha}=\max(G(x,y)_{t}-G(x,y)_{t+1})\cdot{k}$ where $k=0.25$ is a constant determined experimentally. 
   \end{enumerate}

The biggest advantage of the SMR is that pixel differences above $\alpha$ are not contributing to the matching similarity output. These pixels may be outliers or points that dramatically change appearance, and thus should not effect the matching similarity. Outlying pixels usually only increase the error and cause failure,  so we chose to ignore them in this method. This way only reliably matching pixels contribute to the output of each matching step.

\section{\uppercase{Results}}

We tested this approach on a challenging benchmark: the TLD \cite{KalalTLD2010} dataset. From this dataset six videos with different properties were selected as displayed in Table \ref{tab:properties}. Each video contains only one target. The metric used is the number of correctly tracked frames. For this test color videos were converted to grayscale. State-of-the-art performance was achieved and results are presented in Table \ref{tab:example1}.

To illustrate how the qualitatively different way of defining the tracking problem of the SMR tracker provides better results than the traditional approach, we will compare the SMR tracker with the SAD tracker in the present section. 

Figure \ref{fig:example1} shows the detections from the SAD tracker and the SMR tracker where they have used the same template. Points that dramatically changed appearance cause the SAD tracker to fail whereas the SMR tracker correctly detects the object. For illustration purposes, the differences for each pixel between the template and the patches the SAD tracker  and the SMR tracker  detected are mapped in  Figure \ref{fig:example1}. The patch the SMR tracker detected has a bigger sum of absolute differences. However, that is because of the region that dramatically changed appearance. That patch has many close matches with the template as can be seen in Figure \ref{fig:Hist}. As such, the SMR tracker is able to detect it. Again, with the same principle the SMR tracker is able to track the object when it is going out of the scene as shown in Figure \ref{fig:extension}.  

\begin{figure}[!h]
  \centering
   {\epsfig{file = 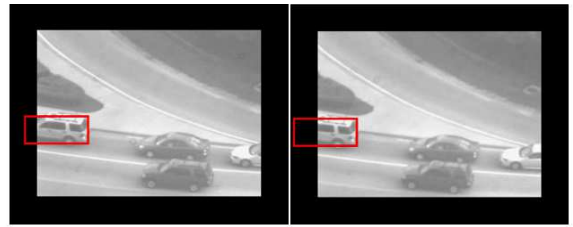, trim =1.4cm 13cm 12cm 4cm, clip ,width = 7.6cm}}
  \caption{The red boxes are the SMR tracker's outputs. The video frame is extended and padded by zeroes. The SMR tracker is able to track when the target is going out of the frame. The template update is ceased in these situations which prevents the drifting from the object.}
  \label{fig:extension}
 \end{figure}

\begin{figure*}[ht]
\centering
{\epsfig{file = 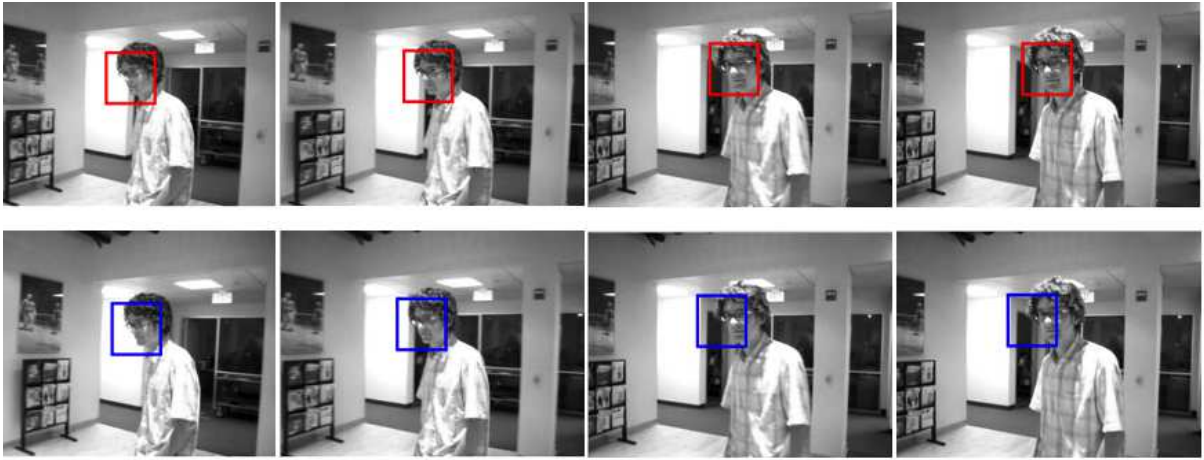, trim =2cm 3.5cm 3cm 11cm, clip , width = 16cm}}
    \caption[Optional caption for list of figures]{(Top) The red boxes are the SMR tracker's outputs. (Bottom) The blue boxes are the SAD tracker's outputs. Outlying pixels cause the SAD tracker to drift, whereas the SMR tracker is not affected by them. } 
   \label{fig:David}
\end{figure*}
 
The SMR tracker is more robust to outliers than the traditional approach. As can be seen in Figure \ref{fig:David} outliers cause the SAD tracker to drift away from the object, whereas the SMR tracker (Figure \ref{fig:David}) finds the target. Ideally the bounding box should be entirely filled with the target. However, during long-term tracking, the object may move back and forth and rotate which cause some background pixels to be included in the next template. A tracker does not know which pixel belong to the object and which ones belong to the background. On the other hand, the SMR tracker has a higher probability of rejecting background pixels, as they tend to change more. 

The SAD tracker from the 2nd frame to 3rd in Figure \ref{fig:David} (bottom) drifts away from the object, because the pixels from the background have become included in the bounding box and they propagate to the template.  When the face moves right, the SAD tracker does not move and drifts away from the object because the background, which has high contrast, gives big differences if the bounding box shifts to a new position. Therefore, the traditional approach gives priority to preventing big distances when it is making a decision, even if these pixels are not the majority of the template. On the other hand, the SMR tracker is focusing on the number of pixels that have small differences with the template which is the face in this case Figure \ref{fig:David} (top).

\section{\uppercase{Failure Mode}}
Even though the SMR tracker updates the template at every frame in this presented work, drifts caused by the accumulation of small errors during each detection are not observed by applying this method on the benchmark dataset. However, when an object becomes occluded very slowly, updating the template at every frame causes the template to include foreground pixels that are not belong to the object. An example can be seen in Figure \ref{fig:Failure}. A better template update mechanism will prevent this kind of failure. This will most probably require the use of a classifier which is out of the scope of the work in this paper. 

\begin{figure}[h]
  \centering
   {\epsfig{file = 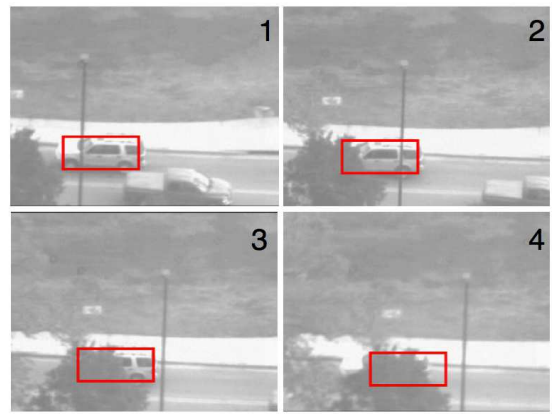, trim =1.4cm 3cm 6cm 1cm, clip ,width = 7.6cm}}
  \caption{ Red boxes are the SMR tracker's results. The every-frame template update causes the outlying pixels to propagate to the templates. When outlying pixels dominate the template, the SMR tracker fails.}
  \label{fig:Failure}
 \end{figure}

\section{\uppercase{Conclusion}}

This paper proposes a novel approach of tracking: the Similarity Matching Ratio (SMR). The SMR tracker is more robust to outliers than the traditional approaches because it is not collecting differences between the template and the frame for each pixel. Instead, it is collecting probabilities from the pixels that have small differences from the template. The SMR tracker tries to find a region which maximizes the good match not minimizes the differences for the whole template. This proves to be a superior approach. The SMR tracker is tested on challenging video sequences and achieves state-of-the-art performance  (See Table \ref{tab:example1}). 

\bibliographystyle{apalike}
{\small
\bibliography{dSMR}}

\vfill
\end{document}